\pdfoutput=1

\documentclass[11pt]{article}

\usepackage{emnlp2021}

\usepackage{times}
\usepackage{latexsym}

\usepackage[T1]{fontenc}

\usepackage[utf8]{inputenc}

\usepackage{microtype}

\usepackage{microtype}
\usepackage{times}  
\usepackage{helvet} 
\usepackage{courier}  
\usepackage{graphicx} 
\usepackage{natbib}  
\usepackage{multirow}
\usepackage{amsmath}
\usepackage{caption} 
\usepackage{diagbox}
\usepackage{arydshln}

\title{A Joint and Domain-Adaptive Approach to Spoken Language Understanding}

\author{Linhao Zhang\textsuperscript{\rm 1}, Yu Shi\textsuperscript{\rm 2}, Linjun Shou\textsuperscript{\rm 2},  Ming Gong\textsuperscript{\rm 2},  Houfeng Wang\textsuperscript{\rm 1}, 
Michael Zeng\textsuperscript{\rm 2} \\ 
\textsuperscript{\rm 1}MOE Key Lab of Computational Linguistics, Peking University\\ 
\textsuperscript{\rm 2}Microsoft\\
\{zhanglinhao, wanghf\}@pku.edu.cn\\
\{yushi, lisho, migon, nzeng\}@microsoft.com
}

\begin{document}

\maketitle

\begin{abstract}


Spoken Language Understanding (SLU) is composed of two subtasks: intent detection (ID) and slot 
filling (SF). There are two lines of research on SLU. One jointly 
tackles these two subtasks to improve their prediction accuracy, and the other focuses on 
the domain-adaptation ability of one of the 
subtasks. In this paper, we attempt to bridge these two lines of research and propose a joint and 
domain adaptive approach to SLU.  
We formulate SLU as a constrained generation task and utilize a dynamic vocabulary based on domain-specific ontology.
We conduct experiments on the ASMixed and MTOD datasets and achieve 
competitive performance with previous state-of-the-art joint models. Besides, results show that our 
joint model can be 
effectively adapted to a new domain.

\end{abstract}

\section{Introduction}
Spoken Language Understanding (SLU) is a critical component in spoken dialogue 
systems. It usually involves two subtasks: intent detection (ID) and slot filling (SF). 
ID aims to identify the intent of the user, while SF aims to
extract the necessary information in the form of slots.

In recent years, there are two lines of research on SLU. One aims to improve the prediction accuracy of 
ID and SF. These models often learn ID and SF jointly by regarding ID as an utterance classification 
problem and SF as a sequence labeling problem. Following this \emph{Classify-Label} framework, various 
joint models have been proposed \cite{Liu2016attention, zhang2016joint, goo2018slot, niu2019novel,zhang2020graph}. These joint models can utilize the semantic correlation between intent and slot and hence result in higher prediction accuracy than separate models.
Despite its success, the \emph{Classify-Label} framework lacks  domain adaptation ability.
This is because the category label spaces of the source domains and target domains, which are made up of class indexes, are not necessarily equivalent.

The other line of research aims to improve models' domain adaptation ability. These models only focus on one of the subtasks (either ID \cite{xia2018zero, liu2019reconstructing, zhang2020discriminative} or SF \cite{bapna2017towards, shah2019robust, liu2020coach}). However, the separate approach has been shown to be inferior to the joint approach in terms of prediction accuracy as it fails to utilize the semantic correlation between slot and intent \cite{zhang2016joint,goo2018slot,zhang2020graph}.

\begin{figure}
\centering
\hspace*{-0.5cm}  
\includegraphics[width=7.5cm]{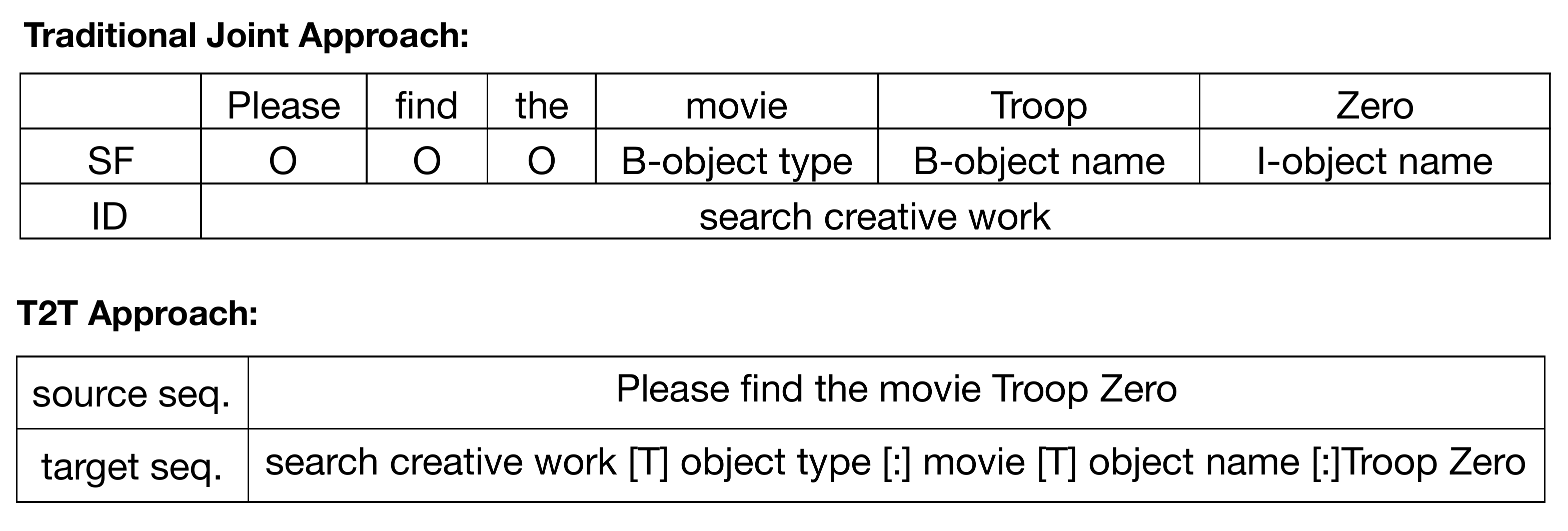}
\caption{Traditional joint approach regards ID as a 
classification task and SF a sequence labeling task, making domain adaptation impossible.  By contrast, T2T approach regards SLU as a generation task, where all ID/SF labels are expressed in natural language in the output sequence.} 
\label{figure-example}
\end{figure}

In this paper, we attempt to bridge these two lines of research and propose a joint and domain-adaptive approach to SLU. Different from previous joint models which follows the  \emph{Classify-Label} framework, we approach SLU from a relatively new perspective by formulating SLU as a text-to-text (T2T) task. As shown in Figure \ref{figure-example}, we define a general format of the output sequence as: \emph{$<$intent$>$ [T] $<$slot name$>$ [:] $<$slot value$>$ [T] $<$slot name$>$ [:] $<$slot value$>$ ... }, where \emph{[T]} and \emph{[:]} are separators. Here all the 
\emph{$<$intent$>$}, \emph{$<$slot name$>$} and \emph{$<$slot value$>$} are expressed in natural language. 

For this T2T setting, a natural model choice is the popular Seq2Seq framework \cite{cho2014learning,sutskever2014sequence} with copy mechanism \cite{see2017get} to tackle this task. Since this model sets no 
constraints to the output, we name it as \emph{Unconstrained-T2T} (UT2T). It depends on the model itself 
to infer the intent and slots using any words in the vocabulary, hence the output may not exactly 
match domain definition, even if the semantic may be correct. 

To this end, we further propose the \emph{Constrained-T2T} (CT2T) that utilizes different 
vocabularies for different segments of the output sequence. It also supports domain-specified intent/slot definitions. At the 
first decoding step, it generates words using the intent vocabulary. To decode the following slot 
name-value pairs, it learns how to alternatively select tokens from the slot vocabulary and the 
input utterance.
For domain adaptation, we feed the 
domain-specified intent/slot names into the model, along with the input utterance. Even if some 
intents/slots are not exactly seen in the training domains, our model can utilize the semantic 
information their names convey to generate the correct results. For example, if our model has seen the 
intent \emph{cancel alarm} in the \emph{Alarm} domain, then it may well be able to generate \emph{
cancel reminder} in the \emph{Reminder} domain. 

We conduct experiments on two major multi-domain SLU datasets, ASMixed and MTOD. Our \emph{CT2T} achieves sentence-level accuracy of 84.87\% and 91.24\% on 
the two datasets, respectively, on par with the best joint model following the \emph{Classify-Label} framework. 
Besides, both few-shot and zero-shot experiments show that it can be effectively adapted 
to a new domain.	 





\section{Related Work}
\label{section-related}

\textbf{Joint Learning. }
In recent years, the \emph{Classify-Label} framework has been the default design for joint ID and SF. It regards ID as an utterance classification task and SF as a sequence labeling task. 
Following this framework, various joint models have been proposed
\cite{Liu2016attention, zhang2016joint, goo2018slot,niu2019novel, qin2019stack, liu2019cm,  zhang2020graph, wu2020slotrefine}. 
Since these joint models can utilize the semantic correlation between
intent and slot, they often result in higher prediction accuracy than separate models.

There are few joint models investigating the problem of domain adaptation. Most of these models 
conduct experiments on single domain datasets such  as ATIS \cite{hemphill1990atis} and 
Snips \cite{Coucke2018SnipsVP}.  There have also been joint models that focused on multi-domain SLU \cite{ liu2017multi, Kim2017ONENETJD}, yet these models are not domain-adaptive. 
Although \citet{qin2020multi} conduct domain adaptation experiments, they actually train a new model using data of both the source and target domains when adapting to the target domain\footnote{Please refer to Section 4.5.4 of their paper}.



\textbf{Domain Adaptation.} There are also SLU models focusing on domain adaptation. 
For example, \citet{bapna2017towards,shah2019robust,liu2020coach} utilize slot descriptions to achieve zero-shot SF. In a sense, their methods are close to our work, as they often use tokenized slot names in place of slot description in practice. However, they need to perform multiple times of labeling for each slot \cite{bapna2017towards, shah2019robust} or require a two-steps pipeline to decide the exact slot types \cite{liu2020coach}, while our model decodes the whole output sequence once and obtains all slot name-value pairs. Besides, there are also 
works that investigate zero/few-shot ID \cite{xia2018zero,liu2019reconstructing,lin2019deep,yan2020unknown,zhang2020discriminative}.

In general, all these works are restricted to either ID or SF, hence cannot enjoy the benefits brought by joint learning. 

\textbf{Seq2Seq for SLU.} Seq2Seq learning was first proposed by \citet{cho2014learning,sutskever2014sequence} for Machine Translation. 
There are previous works that apply the Seq2Seq framework to SLU \cite{Liu2016attention,Zhu2017EncoderdecoderWF}. However, they still follow the \emph{
Classify-Label} framework, meaning that the output of their decoder is a label sequence following the BIO format, rather than natural language. Therefore, they still suffer from the two limitations we mentioned before. 

For SLU, the slot values are all from the source utterance, hence we add the copy mechanism \cite{ Vinyals2015Pointer,see2017get}  into our model. In this respect, our \emph{UT2T}
 is close to \citet{zhao2018improving}, who first applied copy mechanism for slot value prediction. 
However, their model only predicts slot values but not corresponding slot names, which limits its practical applications. By contrast, our model can predict intent, slot names and slot values in a single sequence. 
Our work is notably different from \citet{wu2019transferable}, which employ Seq2Seq for the task of dialogue state tracking (DST). They decode slot value $J$ times independently for all the possible slot names, where $J$ is the number of possible slots. Besides, their model does not involve intent detection.

\begin{figure}
\centering
\hspace*{-0.5cm}  
\includegraphics[width=8cm]{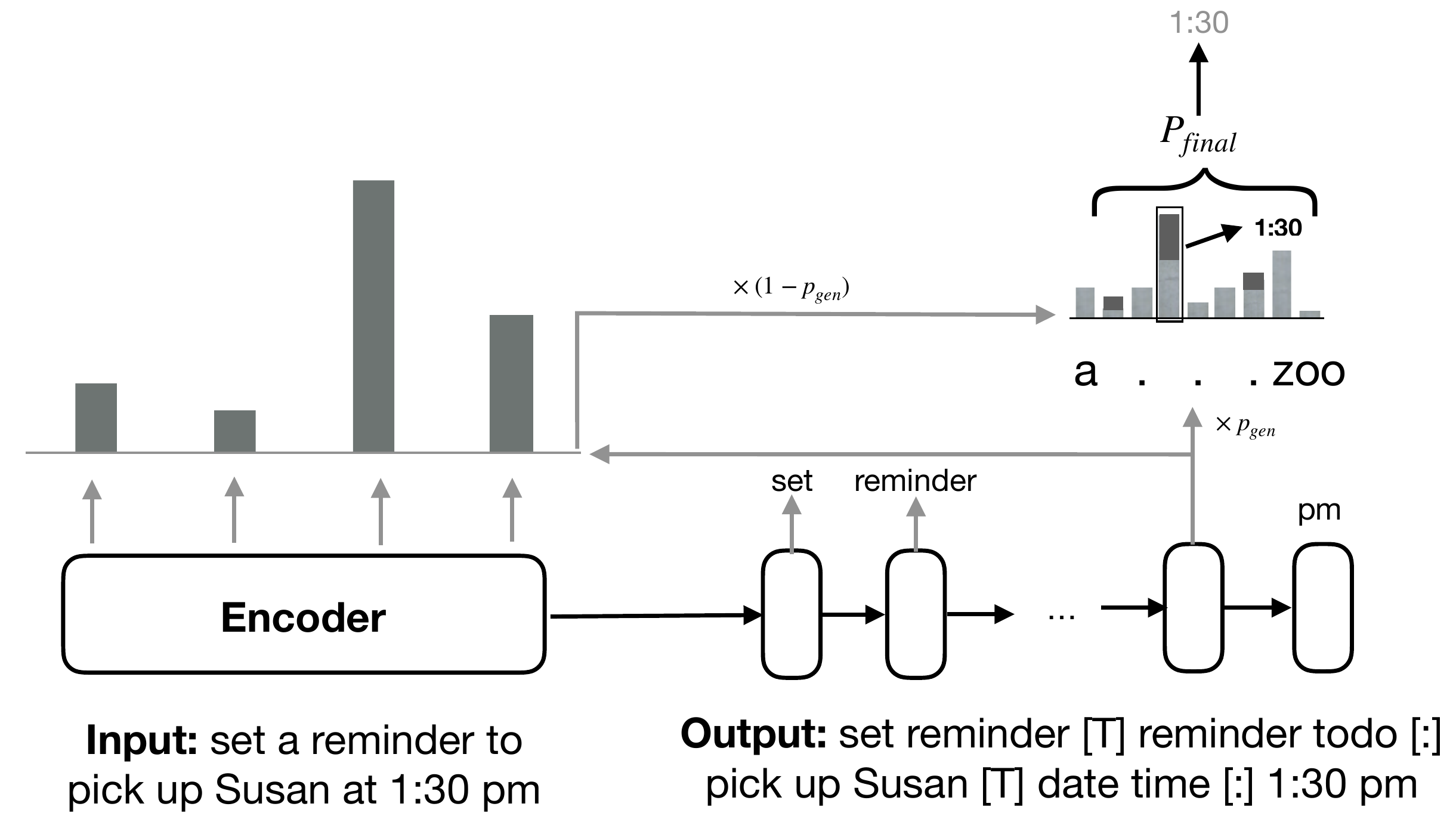}
\caption{\emph{UT2T} is Seq2Seq with copy mechanism without any constraints. It depends on the model itself to infer the intent and slots using any words in the vocabulary.}
\label{figure-model-1}
\end{figure}

\section{Models}

\subsection{Task Formalization}
Given an input utterance $\mathcal{X}$ = ${x_1, x_2, ..., x_n}$, where n denotes the length of the sequence, SF needs to find every slot value in $\mathcal{X}$, and then assign a slot label  to it.  ID aims to decide the intent type of $\mathcal{X}$. 

\begin{figure*}
\centering
\hspace*{-0.5cm}  
\includegraphics[width=15cm]{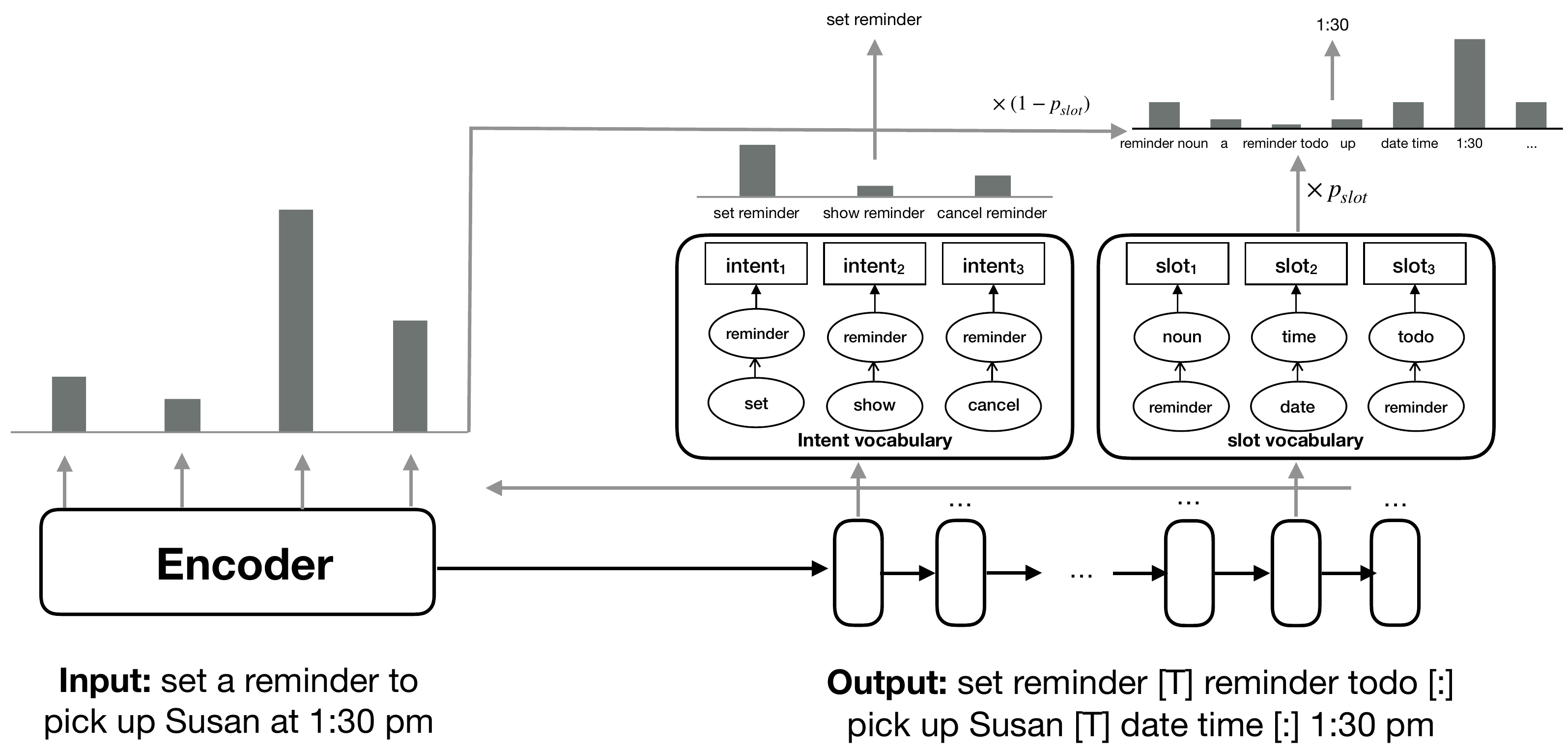}
\caption{Overview of \emph{CT2T}. We first generate the intent using the intent vocabulary. Then, we 
choose between the slot vocabulary and words from the input utterance to generate a series of slot name-value pairs.}
\label{figure-model-2}
\end{figure*}
In this work, we tackle both ID and SF jointly and regard them as a text-to-text task.   We define a general format of the output sequence as: 

\begin{equation}
\begin{aligned}
& \text{$<$intent$>$} [T] \text{$<$slot name$>$} [:] \text{$<$slot value$>$} [T]  \\
& \text{$<$slot name$>$} [:] \text{$<$slot value$>$} ... 
\end{aligned}
\label{equ-output_seq}
\end{equation}

\noindent where \emph{[T]} and \emph{[:]} are separators, and all the 
\emph{$<$intent$>$}, \emph{$<$slot name$>$} and \emph{$<$slot value$>$} are expressed in natural language.  
 
From the tagging-style annotated SLU dataset, this output sequence is constructed based on the following rules:

\begin{enumerate}
\item The intent type of $\mathcal{X}$ is simply put at the beginning of the output sequence, followed by a series of slot name-value pairs.

\item The slot values are extracted from the BIO-tagged sequence. Take Figure \ref{figure-example} as an example, since \emph{Troop} is tagged \emph{B-object name} and \emph{Zero} is tagged \emph{I-object name}, we can hence extract slot value \emph{Troop Zero} and specify its slot name as \emph{object name}.

\item The order of different slot name-value pairs in the output sequence is the same as that of their occurrence in the input utterance. Take Figure \ref{figure-example} as an example, \emph{object type [:] movie} should be  put before \emph{object name [:] Troop Zero} in the output sequence, as \emph{movie} occurs before \emph{Troop Zero} in the input utterance.

\end{enumerate}
 

\subsection{UT2T}
We first explore using a standard 
sequence-to-sequence generation with copy mechanism \cite{see2017get} to tackle this problem. 
As shown in Figure \ref{figure-model-1}, We first encode the input utterance $\mathcal{X}$ = ${x_1, x_2, ..., x_n}$ into  $\mathcal{H}$ = ${h_1, h_2, ..., h_n}$. In this work, we experiment with both non-pretrained LSTM encoder and pretrained RoBERTa \cite{liu2019roberta} encoder.

For the decoder, we use LSTM and update its hidden state $s_t$ at time step t:

\begin{equation}
s_t = LSTM(\overline{e}_t,s_{t-1})
\label{equation-decoder}
\end{equation}
where $\overline{e}_t$ is the embedding of the previous word. While 
training, this is the previous word of 
the ground truth; at test time it is the previous word emitted by the decoder.

The attention distribution is calculated as in \cite{luong2015effective}:

\begin{equation}
\alpha_{i}^{t}= softmax(s_{t} W_{y} h_{i})
\end{equation}
where $W_y$ are model parameters.

Then, the attention weights $\alpha_i^t$ are used to produce a weighted sum of the encoder hidden states, known as the context vector $c_t$, which is then concatenated with the decoder state $s_t$ to produce the vocabulary distribution $P_{vocab}$:

\begin{equation}
\begin{aligned} &c_{t}=\sum_{i} \alpha_{i}^{t} h_{i}
\\ &m_{t} =\tanh \left(W_{a}\left[c_{t} ; s_{t}\right]+b_{a}\right) \\&P_{\text {vocab }} =softmax\left(m_{t}\right) \end{aligned}
\label{equation-attention}
\end{equation}
where \(W_{a}, b_{a}\) are model parameters. 

Note that the slot values are not always in the vocabulary.  To solve this  out-of-vocabulary (OOV) problem, we further employ the copy mechanism \cite{see2017get} to copy slot values from the input utterance.  The attention distribution $\alpha_{i}^t$ and the vocabulary distribution $P_{vocab}$ are then weighted and summed to obtain the final word distribution:
\begin{equation}
p_{\mathrm{gen}}=\sigma\left(w_{a}^{T} m_{t}+w_{s}^{T} s_{t}+w_{e}^{T} \overline{e}_{t}+b_{\mathrm{gen}}\right)
\end{equation}
\begin{equation}
P(w)=p_{\mathrm{gen}} P_{\mathrm{vocab}}(w)+\left(1-p_{\mathrm{gen}}\right) \sum_{i : w_{i}=w} \alpha_{i}^{t}
\end{equation}
where $w_a$, $w_s$, $w_e$, and $b_{gen}$ are model parameters. If w is an OOV word and appears in the source utterance, then $P_{vocab}(w)$ is zero and $\sum_{i : w_{i}=w} \alpha_{i}^{t}$ is not zero. In this way, the model is able to produce OOV words, rather than being restricted to their pre-set vocabulary.

\subsection{CT2T}
\emph{UT2T} depends on the model itself  to infer the intent and slots using any words in the vocabulary, hence the output may 
not exactly match domain definition. Take the example of Figure \ref{figure-example}, the model may generate \emph{find creative work}, rather than \emph{search create work}, failing to match the definition of domain \emph{Reminder}. To solve this limitation, we further propose the \emph{CT2T}.

As mentioned above, 
the output format is defined in Equation \eqref{equ-output_seq}
where \emph{$<$intent$>$} and \emph{$<$slot name$>$} are defined in a domain-specific ontology, and \emph{$<$slot value$>$} is a span of the input sequence. It is a sequence starting with the intent, followed by a series of slot name-value pairs. We can hence exploit this pattern and construct a small, dynamic vocabulary for different segments of the output sequence.

As shown in Figure \ref{figure-model-2}, we first encode the input utterance $\mathcal{X}$ = ${x_1, x_2, ..., x_n}$ into  $\mathcal{H}$ = ${h_1, h_2, ..., h_n}$ with our encoder. In this work, we experiment with both LSTM and pretrained RoBERTa encoders. 

At the first decoding step, we feed the intent vocabulary $\mathcal{I}={intent_1,intent_2, ..., intent_{N_i}}$ to the model, where $N_i$  is the number of domain-specific intents. Each intent $intent_i$ is composed of $T$ words ${w_{i1},w_{i2},...,w_{iT}}$, where $T$ may vary among different intents. We encode $intent_i$ into a fixed-length vector via max-pooling:

\begin{small}
\begin{equation}
vec^{intent}_i = Pooling(Encoder(w_{i1},w_{i2},...,w_{iT}))
\label{equation-intent-vec}
\end{equation}
\end{small}

After obtaining intent vectors $vec^{intent}_i$, we compute the attention scores between the current hidden state $s_1$ and $vec^{intent}_i$:

\begin{equation}
\delta_{i}=softmax\left(s_{1} W_{i} vec^{intent}_i\right)
\label{equation-intent}
\end{equation} 

\noindent where $W_i$ is model parameter.
The intent with the highest attention weight $\delta_{i}$ is outputted. Note that the multi-words intent is outputted in one decoding step.

To decode the following slot name-value pairs, we choose from the slot vocabulary and words from the input utterance. The slot vocabulary $\mathcal{S}={slot_1,slot_2, ..., slot_{N_s}}$ contains $N_s$  possible slots. Each slot $slot_i$ is composed of $T$ words ${w_i1,w_i2,...,w_iT}$, where $T$ may vary among different slots. We encode each slot $slot_i$ into a fixed-length vector:

\begin{small}
\begin{equation}
vec^{slot}_i = Pooling(Encoder(w_{i1},w_{i2},...,w_{iT}))
\label{equation-slot-vec}
\end{equation}
\end{small}

Then we calculate the attention scores $\gamma_{i}^t$ between the current hidden state $s_t$ and slot vector $vec^{slot}_i$, and the attention score $\alpha_i^t$ between $s_t$ and the input hidden state $h_i$:

\begin{equation}
\begin{aligned}
 & \gamma_{i}^t=softmax\left(s_{t} W_{s} vec^{slot}_i\right) \\ 
& \alpha_{i}^t=softmax\left(s_{t} W_{h} h_i\right) 
\end{aligned}
\end{equation} 
where $W_s$, $W_h$ are model parameters. 

This two distributions $\gamma_{i}^t$ and $\alpha_i^t$  are then weighted and summed to obtain the final word distribution.
\begin{equation}
p_{\mathrm{slot}}=\sigma\left(w_{a}^{T} m_{t}+w_{s}^{T} s_{t}+w_{e}^{T} \overline{e}_{t}+b_{\mathrm{slot}}\right)
\end{equation}

\begin{small}
\begin{equation}
P(w)=p_{\mathrm{slot}} \sum_{i : w_{i}=w} \gamma_{i}^{t}+\left(1-p_{\mathrm{slot}}\right) \sum_{i : w_{i}=w} \alpha_{i}^{t}
\end{equation}
\end{small}

where $w_a$, $w_s$, $w_e$ and $b_{slot}$ are model parameters. $m_t$ is calculated as in 
Equation \ref{equation-attention}. The $p_{slot}$ can be seen as a soft switch to choose between slot vocabulary (
for \emph{$<$slot name$>$}) and from the input utterance (for\emph{$<$slot value$>$}). 

Note that the multi-words slot name is outputted as a whole, while slot value is outputted one word at a time.

Unlike traditional joint models, \emph{CT2T} can be transfered to a new domain. Even though there 
may exist intents/slots that are not exactly seen in the training domains,  it can utilize the 
semantic information their names convey to generate the correct ones. \emph{CT2T} also improves on previous separate, domain-adaptive models. This is because \emph{CT2T} is a joint model, the information of one task can be utilized in the other task to promote each other.


\section{Experiments}

\subsection{Datasets}
Following \citet{qin2020multi}, we conducted experiments on the ASMixed and MTOD datasets. The statistics of the two datasets are shown in Table \ref{table-datasets}.

\begin{table}[]
\begin{center}
\scalebox{0.8}{
\begin{tabular}{|l|c|c|}
\hline
 & \textbf{ASMixed}  & \textbf{MTOD}\\ \hline
\# Training & 17,562 & 30,527\\ 
\# Validation   & 1,200 & 4,181  \\ 
\# Test  & 1,593 & 8,621 \\ 
\# Slot          & 192  & 11 \\ 
\# Intent                & 28  & 12  \\ 
\# domains        & ATIS, Snips  & Reminder, Alarm, Weather  \\  \hline
\end{tabular}
}
\end{center}
\caption{\label{table-datasets} Statistics of ASMixed and MTOD datasets.}
\end{table}

\begin{table*}[t!]
\begin{center}
\scalebox{0.8}{
\begin{tabular}{|l|l|l|l|l|l|l|l|}
\hline \multicolumn{1}{|c|}{\multirow{2}*{\textbf{Model}}} & \multicolumn{3}{|c|}{\textbf{ASMixed}}  & \multicolumn{3}{|c|}{\textbf{MTOD}} \\ \cline{2-7}
& \multicolumn{1}{|c|}{ID} & \multicolumn{1}{|c|}{SF} & \multicolumn{1}{|c|}{Sent.} & \multicolumn{1}{|c|}{ID} & \multicolumn{1}{|c|}{SF} & \multicolumn{1}{|c|}{Sent.} \\ \hline
Shared-LSTM \cite{hakkani2016multi}  & 94.41 & 92.55 &76.71& 98.70 & 94.87 &88.71 \\ 
Separate-LSTM \cite{hakkani2016multi}  &94.79 &92.94 &79.53& 99.01& 94.89 &89.73 \\ 
Multi-Domain adv \cite{liu2017multi} &94.79 &92.94 &79.47 &99.01 & 94.89 &88.82 \\ 
One-Net \cite{Kim2017ONENETJD} & 93.72&93.38 &78.28 &98.56 & 95.25 &  89.36\\ 
Local-agnostic-Universal \cite{lee2019locale}& 96.48 & 92.10 &79.35&99.12&94.16 & 88.54\\ 
Domain-Aware $\dagger$ \cite{qin2020multi} &97.30 &94.30 &84.81& 99.20 &95.69 &\textbf{91.27}\\  \hline
UT2T  & 96.74 & 93.37 & 83.55 & 99.11 & 95.48 & 90.99 \\ 
CT2T  & 97.49 &94.34  & \textbf{84.87} & 99.21 & 95.54 & 91.24 \\ \hline
\end{tabular}
}
\end{center}
\caption{\label{table-results} Main results on the ASMixed and MTOD datasets (\%). Best 
sentence-level accuracy results are boldfaced. $\dagger$ means using external knowledge.}
\end{table*}

\textbf{ASMixed} -
The ASMixed \cite{qin2020multi} dataset was created  by mixing the ATIS \cite{hemphill1990atis}
and Snips \cite{Coucke2018SnipsVP} datasets. 
The ATIS \cite{hemphill1990atis} dataset has long been used as a benchmark in SLU. There are 4478 utterances in the training set, 500 in the valid set, and 893 in the test set, with a total of 120 distinct slot labels and 21 different intent types.
The Snips dataset was created by \emph{snips.ai} \cite{Coucke2018SnipsVP}. It is in the domain of personal assistant commands. There are 72 slot labels and 7 intent types. 

\textbf{MTOD} -
The MTOD \cite{schuster2018cross} dataset contains three domains including \emph{alarm}, \emph{reminder}, and \emph{weather}. We follow the same format and partition as in \cite{schuster2018cross,qin2020multi}. There are 30521, 4181, and 8621 utterances in the training, validation, and test set, respectively. There are in total 12 intent types and 11 slot types.

\subsection{Evaluation Metrics}
\label{sect-metrics}
We extract the intent and slots from the output sequence using separator \emph{[T]} and adopt three mainstream evaluation metrics:

We evaluate the system's performance on SF using the F1 score, which is defined as the harmonic average of precision and recall. The metric for ID is  classification accuracy. Besides, following previous work of \citet{goo2018slot,niu2019novel,qin2020multi}, we also report the sentence-level accuracy, which considers both SF and ID performance. A sentence is counted as correct if all its slots and intent are correctly predicted. 


\subsection{Implementation Details}

For both the pretrained and non-pretrained model, we set the batch size to  128. 
Dropout \cite{Hinton2012ImprovingNN} layers are applied on both input and output 
vectors during training for regularization. We use greedy 
decoding for the decoder. 

For the non-pretrained model, we use LSTM as the encoder. The dimensions of LSTM hidden state  and embeddings are both set to 256. We use Adam for the training process to minimize the cross-entropy loss, with learning rate = $10^{-3}$, $\beta_{1}=0.9$, $\beta_{2}=0.98$, and $\epsilon=10^{-9}$.  

For the pretrained model, we employ the pretrained RoBERTa-base model as our encoder\footnote{https://github.com/huggingface/transformers}. The 
dimensions are set to 768. We adopt AdamW \cite{loshchilov2018fixing} as our 
optimizer. Since the encoder makes use of a pretrained model, whereas the decoder 
needs to be trained from scratch, we use different learning rate schemes for the 
encoder and the decoder. We set the peak learning rate and warmup proportion to 4e-5 
and 0.2 for the encoder and 1e-4 and 0.1 for the decoder, respectively.
The token embedding matrix of the decoder is shared with that of RoBERTa.

We use teacher forcing for model training where the ground truth instead of the predicted ones is used. During training, we found that the MTOD dataset is more likely to overfit. We train our model for 50 and 200 epochs, and the dropout rates are set to 0.6 and 0.5 for the ASMixed and MTOD datasets, respectively.
For all the experiments, we select the model which reports the highest sentence-level accuracy on the validation set and evaluate it on the test set.

\begin{figure*}
\centering
\hspace*{-0.5cm}  
\includegraphics[width=14cm]{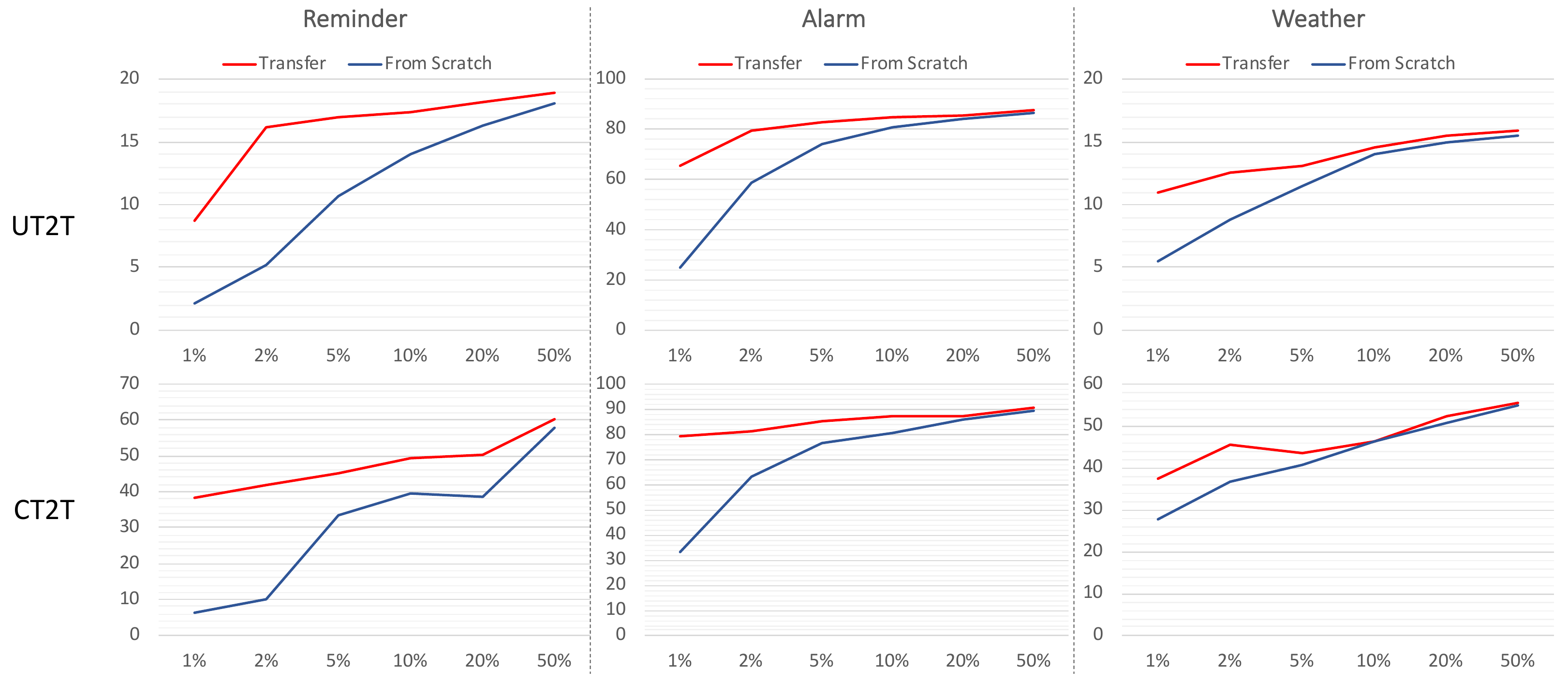}
\caption{\label{figure-few-shot} Domain adaptation results. \emph{Reminder}, \emph{Alarm} and \emph{Weather} stand for the three domains in the MTOD dataset. We report sentence-level accuracy that considers both ID and SF. Note that the comparison is not only between \emph{Transfer} and \emph{From-Scratch}, but also between \emph{UT2T} and \emph{CT2T}.}
\end{figure*}

\subsection{Systems for Comparison}
We compared our model against the following multi-domain SLU baselines\footnote{All the baseline results are taken from \citet{qin2020multi}}:

\textbf{Shared-LSTM} \cite{hakkani2016multi} used a single shared LSTM for both ID and SF for all the domains. 

\textbf{Separated-LSTM} \cite{hakkani2016multi} performed ID and SF for each domain separately. 

\textbf{Multi-Domain Adv} \cite{liu2017multi} proposed an adversarial training model to learn common features that can be shared across multi-domains.

\textbf{One-Net} \cite{Kim2017ONENETJD} jointly performed 
domain, intent, and slot prediction, aiming to alleviate error propagation and lack of 
information sharing.

\textbf{Locale-agnostic-Universal} \cite{lee2019locale} proposed a locale-agnostic 
universal domain classification model that learns a joint representation of an 
utterance over locales with different sets of domains. 

\textbf{Domain-Aware} \cite{qin2020multi} proposed to improve the parameterization of 
multi-domain learning by using domain-specific and task-specific model parameters to 
improve knowledge learning and transfer.



\section{Results}
\subsection{Overall Performance}
\label{sect-overall}

The overall performance on the ASMixed and MTOD datasets are demonstrated in Table \ref{table-results}\footnote{When adopting RoBERTa as encoder, \emph{CT2T} achieves sentence-level accuracy of 87.32\% and 92.19\% on the ASMixed and MTOD and datasets, respectively. However, for a fair comparison, we do not list the results in table \ref{table-results}.}. Note that for a fair comparison with baselines, we assume that we do not know the domain the utterance comes from. To acquire the intent vocabulary for \emph{CT2T}, we extract all the intents from the training data and mix them up to acquire the full intent vocabulary. The slot vocabulary is acquired likewise.

Our first observation is that the \emph{UT2T} performs surprisingly well, given the simplicity of its structure. Specifically, it achieves 83.55\% and 90.99\% sentence-level accuracy on the ASMixed and MTOD datasets, respectively. Although the results are lower than the current state-of-the-art joint model, it demonstrates the feasibility of formulating SLU as a generation task. 

On the other hand, \emph{CT2T}'s performance is on par with the state-of-the-art joint model. 
Specifically, it achieves 97.49\% and 99.21\% intent accuracy on the ASMixed and MTOD 
datasets, respectively, outperforming previous state-of-the-art on both datasets. 
Besides, it results in 94.34\% and 95.54\% slot F1 scores on the two datasets, which 
are on par with \citet{qin2020multi}. Note that \citet{qin2020multi} uses Stanford 
CoreNLP to generate the dependency tree and utilizes it as external knowledge when 
encoding input utterance, while our \emph{CT2T} relies on no additional information beyond 
the dataset.  

Compared with traditional joint models, the reason why \emph{CT2T} gives such competitive performance is that it makes better use of the semantic information of each individual word in intent/slot, rather than regarding intent/slot as class index.


\subsection{Domain Adaptation}
In this section we test our model's domain adaptation ability.
For the \emph{Transfer} setting, each model is trained on two domains of the MTOD dataset, and a held-out domain is reserved. Then we fine-tune our model on the held-out domain with x\% training data and evaluate its performance on the held-out domain. For the \emph{From-Scratch} setting, we omit the training process and directly fine-tune and test the model on the held-out domain\footnote{Note that we do not compare our model with previous separate, domain adaptive models such as \cite{ bapna2017towards,xia2018zero,zhang2020discriminative} for the following reasons: 1) They only focus on one of the subtasks (either ID or SF), while the reported sentence-level accuracy considers both ID and SF. Focusing on only one of the subtasks making the metric meaningless. 2) Comparing only one of the subtasks is also difficult because they do not experiment with the MTOD datasets and many of these papers do not release their source code. Besides, the transfer settings differ from paper to paper. 3)The point of this subsection is to demonstrate the domain-adaptation ability of our joint model, rather than to declare a new state-of-the-art results on ID or SF.}. 
The results are  shown in Figure \ref{figure-few-shot}.

We first note that models following the \emph{Classify-Label} framework cannot 
achieve domain adaptation on any of these domains. This is because the category label 
spaces of the training domains and the held-out domain are not equivalent. 

By formulating SLU as a text-to-text task, \emph{UT2T} demonstrates certain domain adaptation ability. We can see that the \emph{Transfer} curves are higher than the \emph{From Scratch} curves on all three domains, showing the benefits brought by transfer learning. However, we also note that the absolute value  is low, especially for domain \emph{Reminder} and \emph{Weather}. Even with 50\% training data, its sentence-level accuracy on the two domains are lower than 20\%. Detailed analysis shows that this is because \emph{UT2T} does not support user-specified ontology. For example, it can never generate the slot name \emph{Reminder todo}, when trained on the \emph{Alarm} and \emph{Weather} domains and  transfered to the \emph{Reminder} domain, since the word \emph{todo} is in neither the model vocabulary nor the input utterance. 

On the other hand, our \emph{CT2T} gives much more satisfactory performance. Not only its \emph{Transfer} results are higher than the \emph{From Scratch} results on all three domains, but also it achieves much higher absolute value than \emph{UT2T}. For example, it achieves 38.31\% sentence-level accuracy after fine-tuning using only 1\% of \emph{reminder} domain data, outperforming the From-Scratch method by as large as 31.99\%. Besides, it also outperforms the \emph{Transfer} result of  \emph{UT2T} by more than 30\%.


\subsection{Zero-Shot Analysis}
\begin{figure}
\centering
\hspace*{-0.5cm}  
\includegraphics[width=7cm]{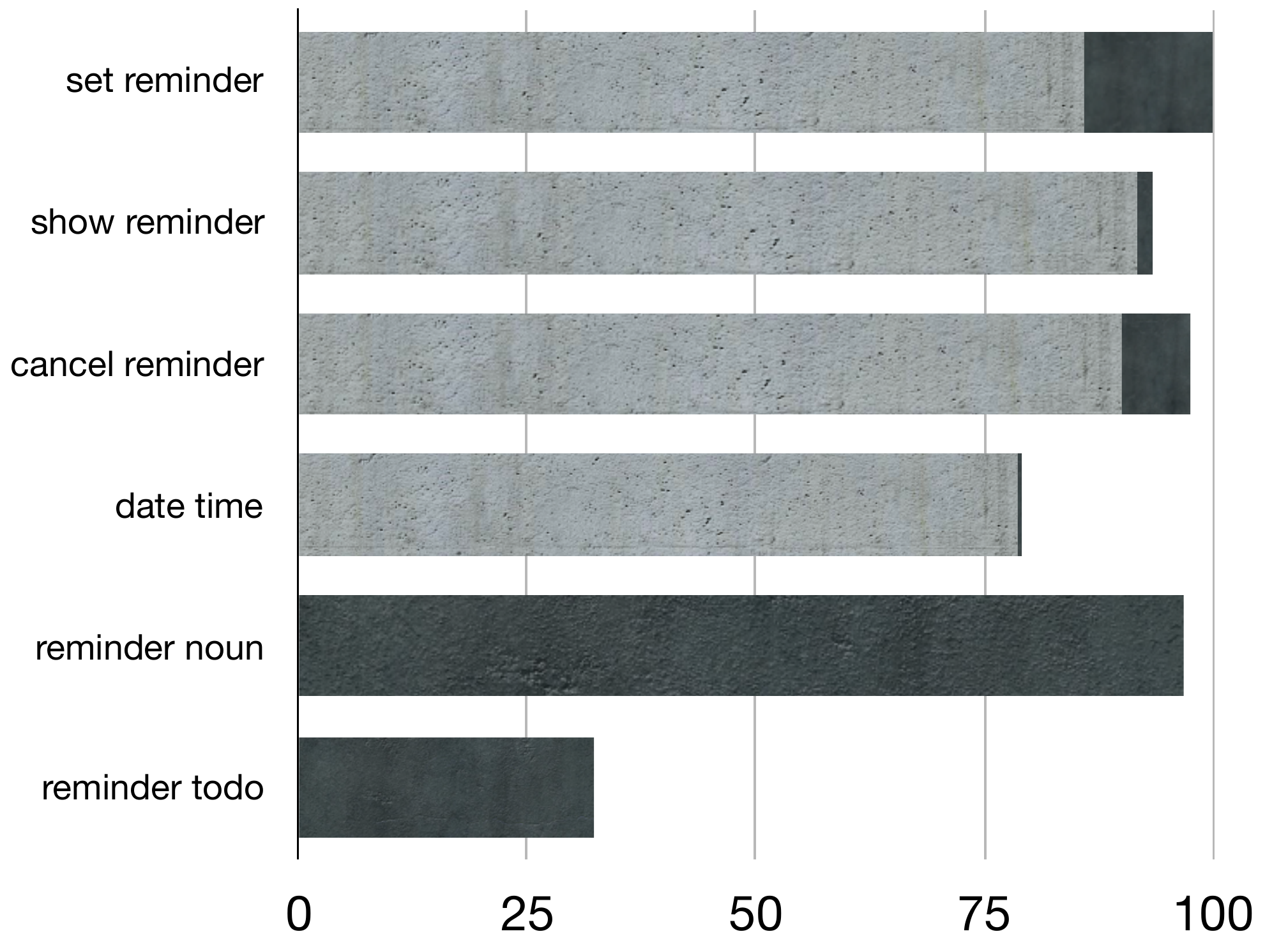}
\caption{Detailed domain adaptation analysis on the \emph{reminder} domain. The first and last three rows stand for intents and slots, respectively. The gray and black bars indicate zero-shot and few-shot (with 1\% training data) results.}
\label{figure-zero-shot}
\end{figure}

We further give detailed analysis on \emph{CT2T}'s zero-shot ability.
We select the \emph{reminder} domain and report performance on each individual intent/slot. The model is first trained on the \emph{alarm} and \emph{weather} domains and then tested on the \emph{reminder} domain without further model parameter update. The results are shown in Figure \ref{figure-zero-shot}.

As we can see, \emph{CT2T} achieves striking accuracy in terms of zero-shot ID on the 
\emph{reminder} domain. The three intents achieve more than 90\% accuracy, without any
training instance of the \emph{reminder} domain. The reason is that although intents such as \emph{cancel reminder} are not seen during training, there are similar intents such as \emph{cancel alarm} in the \emph{alarm} domain. Since \emph{CT2T} regards intents as natural language, the semantic meaning of the word \emph{cancel} is successfully transferred to a new domain, and help our model to generate the correct intent. Note that for the traditional joint model, where the intents are regarded as class indexes, this kind of transfer cannot be realized.

Based on the slots already learned, our model is able to directly track those slots that are present in a new domain. For example, \emph{CT2T} achieves high performance on the \emph{date time} slot on domain \emph{reminder}, as \emph{date time} also appears in the \emph{weather} domain. On the other hand, the zero-shot results on the \emph{reminder noun} and \emph{reminder todo} slots are pool, as the model has never seen similar semantics in the training domains. However, with as little as 1\% training data, the results on these two slots are dramatically improved. 
Another way to solve this problem is to enlarge slot semantics coverage in training data by adding more domains. We leave this to our future work.

\subsection{Case Study}
\begin{figure}
\centering
\hspace*{-0.5cm}  
\includegraphics[width=8cm]{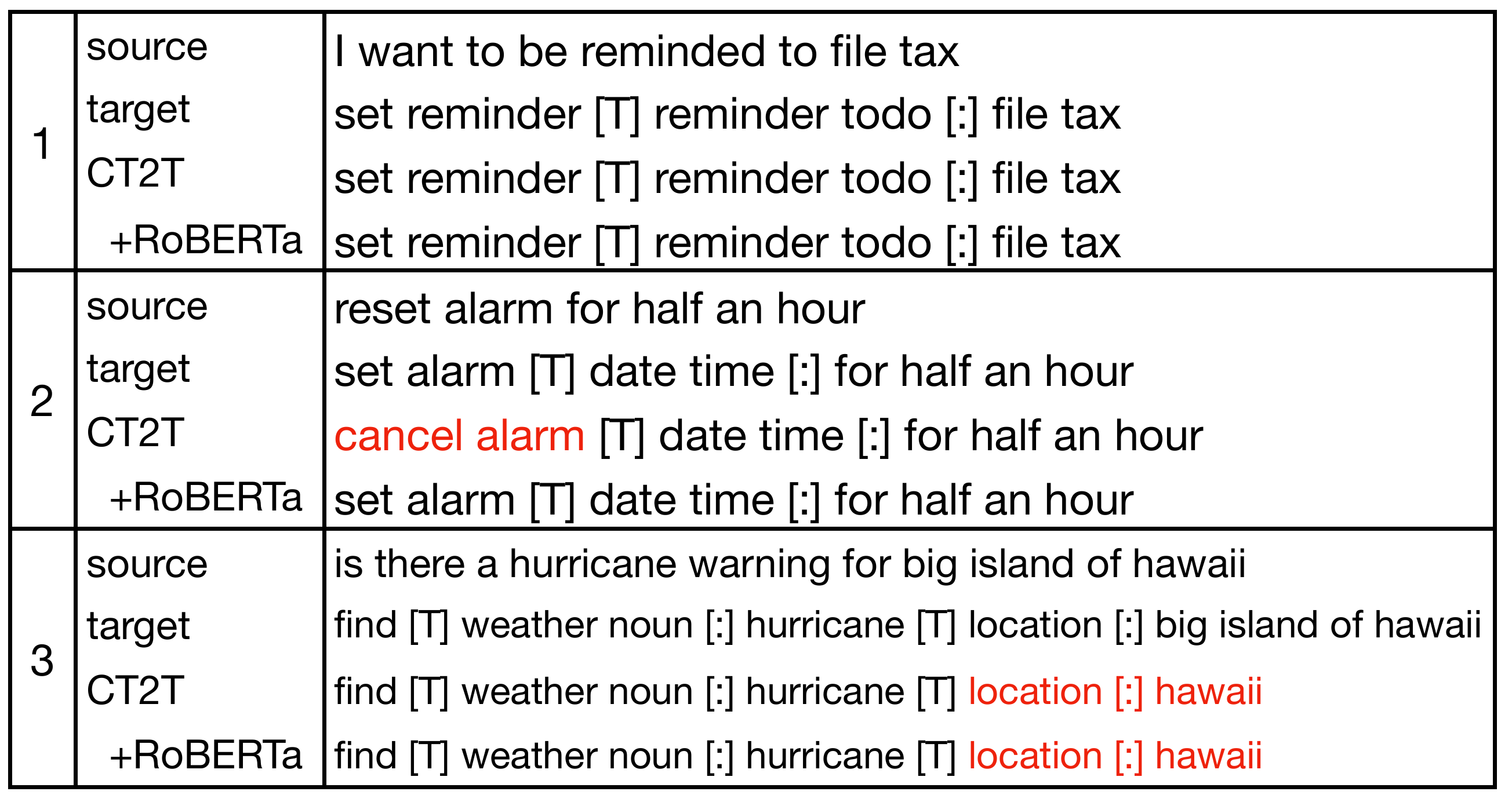}
\caption{Case study.} 

\label{figure-case}
\end{figure}

To better understand the model performance, we provide a case study in Figure \ref{figure-case}.

Our first observation is that the format of the output sequence is well-learned. After 
the model fully converged, we see no cases that break the format rule specified in  Equation \ref{equ-output_seq}, which makes T2T approach to SLU possible.

The second observation is that the pretrained language model is a great help on general semantics.
Take the second case in Figure \ref{figure-case}, the intent of the utterance should be \emph{set alarm}, yet the 
model wrongly predicts it as \emph{cancel alarm}. The reason is that the semantic 
meaning of the word \emph{reset} is not well trained.  When adopting RoEBRTa as the encoder, 
the model is able to fix this type of error. 

We also observe a common error rising from the boundary of slot values. As shown in the third case, the value for slot \emph{location} should be \emph{big island of hawaii}, yet our model simply predicts it as \emph{hawaii}. Even with RoBERTa as our encoder, this error is still not fixed. Strictly speaking, this kind of error is not caused by poor language understanding ability, but the existence of nested entities. There are some works \cite{zheng2019boundary} that aim to solve this problem, but it is beyond the scope of this paper.

\section{Conclusions}
In this paper, we propose a joint and domain adaptive SLU model based on T2T setting. 
We first explore the unconstrained generation approach and show that it is 
workable for SLU. Then, we propose the \emph{CT2T}  where different 
vocabularies are constructed for different segments of the output sequence. Our \emph{CT2T} 
achieves very competitive performance on two SLU datasets. Further experiments demonstrate that 
the model trained on the source domains can be effectively adapted to a new domain.

\bibliography{anthology, custom}

\begin{thebibliography}{36}
\expandafter\ifx\csname natexlab\endcsname\relax\def\natexlab#1{#1}\fi

\bibitem[{Bapna et~al.(2017)Bapna, Tur, Hakkani-Tur, and
  Heck}]{bapna2017towards}
Ankur Bapna, Gokhan Tur, Dilek Hakkani-Tur, and Larry Heck. 2017.
\newblock Towards zero-shot frame semantic parsing for domain scaling.

\bibitem[{Cho et~al.(2014)Cho, Van~Merri{\"e}nboer, Gulcehre, Bahdanau,
  Bougares, Schwenk, and Bengio}]{cho2014learning}
Kyunghyun Cho, Bart Van~Merri{\"e}nboer, Caglar Gulcehre, Dzmitry Bahdanau,
  Fethi Bougares, Holger Schwenk, and Yoshua Bengio. 2014.
\newblock Learning phrase representations using rnn encoder-decoder for
  statistical machine translation.

\bibitem[{Coucke et~al.(2018)Coucke, Saade, Ball, Bluche, Caulier, Leroy,
  Doumouro, Gisselbrecht, Caltagirone, Lavril, Primet, and
  Dureau}]{Coucke2018SnipsVP}
Alice Coucke, Alaa Saade, Adrien Ball, Th{\'e}odore Bluche, Alexandre Caulier,
  David Leroy, Cl{\'e}ment Doumouro, Thibault Gisselbrecht, Francesco
  Caltagirone, Thibaut Lavril, Ma{\"e}l Primet, and Joseph Dureau. 2018.
\newblock Snips voice platform: an embedded spoken language understanding
  system for private-by-design voice interfaces.
\newblock volume abs/1805.10190.

\bibitem[{Goo et~al.(2018)Goo, Gao, Hsu, Huo, Chen, Hsu, and
  Chen}]{goo2018slot}
Chih-Wen Goo, Guang Gao, Yun-Kai Hsu, Chih-Li Huo, Tsung-Chieh Chen, Keng-Wei
  Hsu, and Yun-Nung Chen. 2018.
\newblock Slot-gated modeling for joint slot filling and intent prediction.
\newblock In \emph{NAACL-HLT}, volume~2, pages 753--757.

\bibitem[{Hakkani-T{\"u}r et~al.(2016)Hakkani-T{\"u}r, T{\"u}r, Celikyilmaz,
  Chen, Gao, Deng, and Wang}]{hakkani2016multi}
Dilek Hakkani-T{\"u}r, G{\"o}khan T{\"u}r, Asli Celikyilmaz, Yun-Nung Chen,
  Jianfeng Gao, Li~Deng, and Ye-Yi Wang. 2016.
\newblock Multi-domain joint semantic frame parsing using bi-directional
  rnn-lstm.
\newblock In \emph{Interspeech}, pages 715--719.

\bibitem[{Hemphill et~al.(1990)Hemphill, Godfrey, and
  Doddington}]{hemphill1990atis}
Charles~T. Hemphill, John~J. Godfrey, and George~R. Doddington. 1990.
\newblock The atis spoken language systems pilot corpus.
\newblock In \emph{Speech and Natural Language: Proceedings of a Workshop Held
  at Hidden Valley, Pennsylvania, June 24-27,1990}.

\bibitem[{Hinton et~al.(2012)Hinton, Srivastava, Krizhevsky, Sutskever, and
  Salakhutdinov}]{Hinton2012ImprovingNN}
Geoffrey~E. Hinton, Nitish Srivastava, Alex Krizhevsky, Ilya Sutskever, and
  Ruslan~R. Salakhutdinov. 2012.
\newblock Improving neural networks by preventing co-adaptation of feature
  detectors.
\newblock volume abs/1207.0580.

\bibitem[{Kim et~al.(2017)Kim, Lee, and Stratos}]{Kim2017ONENETJD}
Young-Bum Kim, Sungjin Lee, and Karl Stratos. 2017.
\newblock Onenet: Joint domain, intent, slot prediction for spoken language
  understanding.
\newblock pages 547--553.

\bibitem[{Lee et~al.(2019)Lee, Sarikaya, and Kim}]{lee2019locale}
Jihwan Lee, Ruhi Sarikaya, and Young-Bum Kim. 2019.
\newblock Locale-agnostic universal domain classification model in spoken
  language understanding.

\bibitem[{Lin and Xu(2019)}]{lin2019deep}
Ting-En Lin and Hua Xu. 2019.
\newblock Deep unknown intent detection with margin loss.
\newblock \emph{arXiv preprint arXiv:1906.00434}.

\bibitem[{Liu and Lane(2016)}]{Liu2016attention}
Bing Liu and Ian Lane. 2016.
\newblock Attention-based recurrent neural network models for joint intent
  detection and slot filling.

\bibitem[{Liu and Lane(2017)}]{liu2017multi}
Bing Liu and Ian Lane. 2017.
\newblock Multi-domain adversarial learning for slot filling in spoken language
  understanding.

\bibitem[{Liu et~al.(2019{\natexlab{a}})Liu, Zhang, Fan, Fu, Li, Wu, and
  Lam}]{liu2019reconstructing}
Han Liu, Xiaotong Zhang, Lu~Fan, Xuandi Fu, Qimai Li, Xiao-Ming Wu, and
  Albert~YS Lam. 2019{\natexlab{a}}.
\newblock Reconstructing capsule networks for zero-shot intent classification.
\newblock In \emph{Proceedings of the 2019 Conference on Empirical Methods in
  Natural Language Processing and the 9th International Joint Conference on
  Natural Language Processing (EMNLP-IJCNLP)}, pages 4801--4811.

\bibitem[{Liu et~al.(2019{\natexlab{b}})Liu, Meng, Zhang, Zhou, Chen, and
  Xu}]{liu2019cm}
Yijin Liu, Fandong Meng, Jinchao Zhang, Jie Zhou, Yufeng Chen, and Jinan Xu.
  2019{\natexlab{b}}.
\newblock Cm-net: A novel collaborative memory network for spoken language
  understanding.

\bibitem[{Liu et~al.(2019{\natexlab{c}})Liu, Ott, Goyal, Du, Joshi, Chen, Levy,
  Lewis, Zettlemoyer, and Stoyanov}]{liu2019roberta}
Yinhan Liu, Myle Ott, Naman Goyal, Jingfei Du, Mandar Joshi, Danqi Chen, Omer
  Levy, Mike Lewis, Luke Zettlemoyer, and Veselin Stoyanov. 2019{\natexlab{c}}.
\newblock Roberta: A robustly optimized bert pretraining approach.

\bibitem[{Liu et~al.(2020)Liu, Winata, Xu, and Fung}]{liu2020coach}
Zihan Liu, Genta~Indra Winata, Peng Xu, and Pascale Fung. 2020.
\newblock Coach: A coarse-to-fine approach for cross-domain slot filling.

\bibitem[{Loshchilov and Hutter(2018)}]{loshchilov2018fixing}
Ilya Loshchilov and Frank Hutter. 2018.
\newblock Fixing weight decay regularization in adam.

\bibitem[{Luong et~al.(2015)Luong, Pham, and Manning}]{luong2015effective}
Minh-Thang Luong, Hieu Pham, and Christopher~D Manning. 2015.
\newblock Effective approaches to attention-based neural machine translation.

\bibitem[{Niu et~al.(2019)Niu, Haihong, Chen, and Song}]{niu2019novel}
Peiqing Niu, E~Haihong, Zhongfu Chen, and Meina Song. 2019.
\newblock A novel bi-directional interrelated model for joint intent detection
  and slot filling.
\newblock In \emph{Proceedings of the 57th Conference of the Association for
  Computational Linguistics}, pages 5467--5471.

\bibitem[{Qin et~al.(2019)Qin, Che, Li, Wen, and Liu}]{qin2019stack}
Libo Qin, Wanxiang Che, Yangming Li, Haoyang Wen, and Ting Liu. 2019.
\newblock A stack-propagation framework with token-level intent detection for
  spoken language understanding.

\bibitem[{Qin et~al.(2020)Qin, Ni, Zhang, Che, Li, and Liu}]{qin2020multi}
Libo Qin, Minheng Ni, Yue Zhang, Wanxiang Che, Yangming Li, and Ting Liu. 2020.
\newblock Multi-domain spoken language understanding using domain-and
  task-aware parameterization.

\bibitem[{Schuster et~al.(2018)Schuster, Gupta, Shah, and
  Lewis}]{schuster2018cross}
Sebastian Schuster, Sonal Gupta, Rushin Shah, and Mike Lewis. 2018.
\newblock Cross-lingual transfer learning for multilingual task oriented
  dialog.

\bibitem[{See et~al.(2017)See, Liu, and Manning}]{see2017get}
Abigail See, Peter~J Liu, and Christopher~D Manning. 2017.
\newblock Get to the point: Summarization with pointer-generator networks.

\bibitem[{Shah et~al.(2019)Shah, Gupta, Fayazi, and
  Hakkani-Tur}]{shah2019robust}
Darsh~J Shah, Raghav Gupta, Amir~A Fayazi, and Dilek Hakkani-Tur. 2019.
\newblock Robust zero-shot cross-domain slot filling with example values.

\bibitem[{Sutskever et~al.(2014)Sutskever, Vinyals, and
  Le}]{sutskever2014sequence}
Ilya Sutskever, Oriol Vinyals, and Quoc~V Le. 2014.
\newblock Sequence to sequence learning with neural networks.
\newblock In \emph{Advances in neural information processing systems}, pages
  3104--3112.

\bibitem[{Vinyals et~al.(2015)Vinyals, Fortunato, and
  Jaitly}]{Vinyals2015Pointer}
Oriol Vinyals, Meire Fortunato, and Navdeep Jaitly. 2015.
\newblock Pointer networks.
\newblock In \emph{Advances in neural information processing systems}, pages
  2692--2700.

\bibitem[{Wu et~al.(2019)Wu, Madotto, Hosseini-Asl, Xiong, Socher, and
  Fung}]{wu2019transferable}
Chien-Sheng Wu, Andrea Madotto, Ehsan Hosseini-Asl, Caiming Xiong, Richard
  Socher, and Pascale Fung. 2019.
\newblock Transferable multi-domain state generator for task-oriented dialogue
  systems.

\bibitem[{Wu et~al.(2020)Wu, Ding, Lu, and Xie}]{wu2020slotrefine}
Di~Wu, Liang Ding, Fan Lu, and Jian Xie. 2020.
\newblock Slotrefine: A fast non-autoregressive model for joint intent
  detection and slot filling.

\bibitem[{Xia et~al.(2018)Xia, Zhang, Yan, Chang, and Yu}]{xia2018zero}
Congying Xia, Chenwei Zhang, Xiaohui Yan, Yi~Chang, and Philip~S Yu. 2018.
\newblock Zero-shot user intent detection via capsule neural networks.

\bibitem[{Yan et~al.(2020)Yan, Fan, Li, Liu, Zhang, Wu, and
  Lam}]{yan2020unknown}
Guangfeng Yan, Lu~Fan, Qimai Li, Han Liu, Xiaotong Zhang, Xiao-Ming Wu, and
  Albert~YS Lam. 2020.
\newblock Unknown intent detection using gaussian mixture model with an
  application to zero-shot intent classification.
\newblock In \emph{Proceedings of the 58th Annual Meeting of the Association
  for Computational Linguistics}, pages 1050--1060.

\bibitem[{Zhang et~al.(2020{\natexlab{a}})Zhang, Hashimoto, Liu, Wu, Wan, Yu,
  Socher, and Xiong}]{zhang2020discriminative}
Jian-Guo Zhang, Kazuma Hashimoto, Wenhao Liu, Chien-Sheng Wu, Yao Wan, Philip~S
  Yu, Richard Socher, and Caiming Xiong. 2020{\natexlab{a}}.
\newblock Discriminative nearest neighbor few-shot intent detection by
  transferring natural language inference.

\bibitem[{Zhang et~al.(2020{\natexlab{b}})Zhang, Ma, Zhang, Yan, and
  Wang}]{zhang2020graph}
Linhao Zhang, Dehong Ma, Xiaodong Zhang, Xiaohui Yan, and Houfeng Wang.
  2020{\natexlab{b}}.
\newblock Graph lstm with context-gated mechanism for spoken language
  understanding.
\newblock In \emph{AAAI}, pages 9539--9546.

\bibitem[{Zhang and Wang(2016)}]{zhang2016joint}
Xiaodong Zhang and Houfeng Wang. 2016.
\newblock A joint model of intent determination and slot filling for spoken
  language understanding.
\newblock In \emph{IJCAI}, pages 2993--2999.

\bibitem[{Zhao and Feng(2018)}]{zhao2018improving}
Lin Zhao and Zhe Feng. 2018.
\newblock Improving slot filling in spoken language understanding with joint
  pointer and attention.
\newblock In \emph{Proceedings of the 56th Annual Meeting of the Association
  for Computational Linguistics (Volume 2: Short Papers)}, pages 426--431.

\bibitem[{Zheng et~al.(2019)Zheng, Cai, Xu, Leung, and Xu}]{zheng2019boundary}
Changmeng Zheng, Yi~Cai, Jingyun Xu, Ho-fung Leung, and Guandong Xu. 2019.
\newblock A boundary-aware neural model for nested named entity recognition.
\newblock In \emph{Proceedings of the 2019 Conference on Empirical Methods in
  Natural Language Processing and the 9th International Joint Conference on
  Natural Language Processing (EMNLP-IJCNLP)}, pages 357--366.

\bibitem[{Zhu and Yu(2017)}]{Zhu2017EncoderdecoderWF}
Su~Zhu and K.~Yu. 2017.
\newblock Encoder-decoder with focus-mechanism for sequence labelling based
  spoken language understanding.
\newblock pages 5675--5679.

\end{thebibliography}

\bibliographystyle{acl_natbib}

\end{document}